\documentclass{article}

\usepackage{PRIMEarxiv}

\usepackage[utf8]{inputenc} 
\usepackage[T1]{fontenc}    
\usepackage{hyperref}       
\usepackage{url}            
\usepackage{booktabs}       
\usepackage{amsfonts}       
\usepackage{nicefrac}       
\usepackage{microtype}      
\usepackage{lipsum}
\usepackage{fancyhdr}       
\usepackage{graphicx}       
\graphicspath{{media/}}     

\usepackage{algorithm}
\usepackage{algpseudocode}

\title{Doubly Stochastic Matrix Models\\for Estimation of Distribution Algorithms}

\author{
  Valentino Santucci \\
  University for Foreigners of Perugia \\
  Perugia, Italy\\
  \texttt{valentino.santucci@unistrapg.it} \\
   \And
  Josu Ceberio \\
  University of the Basque Country \\
  San Sebastian, Spain\\
  \texttt{josu.ceberio@ehu.eus} \\
}

\begin{document}
\maketitle

\begin{abstract}
Problems with solutions represented by permutations are very prominent in combinatorial optimization. Thus, in recent decades, a number of evolutionary algorithms have been proposed to solve them, and among them, those based on probability models have received much attention. In that sense, most efforts have focused on introducing algorithms that are suited for solving ordering/ranking nature problems. However, when it comes to proposing probability-based evolutionary algorithms for assignment problems, the works have not gone beyond proposing simple and in most cases univariate models. In this paper, we explore the use of Doubly Stochastic Matrices (DSM) for optimizing matching and assignment nature permutation problems. To that end, we explore some learning and sampling methods to efficiently incorporate DSMs within the picture of evolutionary algorithms. Specifically, we adopt the framework of estimation of distribution algorithms and compare DSMs to some existing proposals for permutation problems. Conducted preliminary experiments on instances of the quadratic assignment problem validate this line of research and show that DSMs may obtain very competitive results, while computational cost issues still need to be further investigated.
\end{abstract}


\section{Introduction}\label{sec:intro}

Permutation problems have been a prominent research topic for the combinatorial optimization community. Although permutation problems belong to the family of combinatorial problems, the developments that have been made over the years have raised specific proposals for this type of problems due to the particularity that distinguishes them: the permutation codification. As stated by~\cite{Stanley1986}, permutations are probably among the richest combinatorial structures. Motivated principally by their versatility, permutations appear in a vast range of domains, such as graph theory, mathematical psychology or bioinformatics, but particularly, in logistic problems such as routing~\cite{toth2001}, scheduling~\cite{Gupta2006} or assignment~\cite{koopmans1955}.

The literature presents a significant number of different permutation problems, however, it is possible to classify them based on the nature of what they represent. Santucci et al.~\cite{Santucci2022}, divide permutation problems into ordering and matching problems. The first class of problems aims to find an ordering/ranking of a given set of items, while the second pretends to match two given equally sized sets of items. It becomes obvious that, due to the different meaning that the permutations have in each case, those problems require different algorithmic approaches. 

With the aim of solving permutation problems, the community of evolutionary computation has proposed many approaches, and has focused with special intensity in the development of probability-based strategies (on a variety of paradigms).

To enumerate some, Ceberio et al. investigated the usage of probability models that describe a probability distribution over permutations such as Plackett-Luce~\cite{ceberio2013plackett} and Mallows models under Kendall's-$\tau$~\cite{ceberio2013a} and Ulam~\cite{irurozki2014e} distances, and evaluated their performance in the framework of Estimation of Distribution Algorithms (EDAs). Ayodele et al.~\cite{Ayodele2016} proposed the use of a transformation between real-coded vectors and permutations called  "Random Key" and approached the permutation flowshop scheduling problem with an EDA. Santucci et al.~\cite{santucci2020gradient} studied model-based gradient search under the Plackett-Luce model for optimizing the linear ordering problem.

A deep literature analysis reveals that when approaching permutation problems, the authors have focused on designing strategies that are compatible with the permutation codification~\cite{baioletti2020variable,guijt2022impact}. However, it becomes obvious to think that there is no model that works well for any permutation problem. In fact, we observe that the models enumerated above focus principally on modelling the ordering/rankings of the items in the permutations, and models that capture the assignment nature of  the problems are less frequent. In this trend, we find the work by Irurozki et al.~\cite{irurozki2018sampling} that developed a Mallows model under the Cayley distance and reported to be suited for modelling permutations from the second class of problems. 
A similar consideration about the suitability of the so-called "exchange" or "interchange" moves for assignment/matching problems is provided also in~\cite{baioletti2019search} and~\cite{baioletti2020experimental}.

In this paper, we aim to continue 
this trend
and propose probability models that are suited to model permutations from matching problems, and then use such models to develop probability-based evolutionary algorithms. With this purpose, we study the use of Doubly Stochastic Matrices (DSMs) as probability models over the space of permutations. It is already known that a permutation can be represented as a 0/1 matrix where each row and column have exactly one 1-entry (and 0 for the rest of the entries). This makes them particular cases of DSMs that relax the constraints to “each row/column are formed by positive real numbers summing-up to 1”. These properties (that will be made clear in Section \ref{sec:dsm}) allows seeing the DSMs as synthesis of multiple permutations, thus making it possible to learn a DSM from permutations and to sample permutations from a DSM. 
DSMs specify the probability of an item appearing at a particular position and, intuitively, they look good models for matching/assignment problems like the well known Quadratic Assignment Problem (QAP)~\cite{koopmans1955}.

On the basis of previous literature on DSMs, mostly exogenous to evolutionary computation community, we investigate 
learning and sampling procedures and analyze their basic theoretical properties. Then, a number of EDA proposals using DSMs as a model are built. Finally, experiments on a selected set of QAP benchmark instances are carried out and the results of the different strategies proposed are compared among them and with respect to the most relevant EDA proposals for the QAP. Moreover, also empirical analysis of convergence are investigated. Results reveal good potential in the usage of DSMs to deal with permutation matching problems.

The remainder of the paper is organized as follows. Section~\ref{sec:perm} provides background on permutation problems, the encodings used and their nature. It also presents a distinction on the classes of problems. Next, in Section~\ref{sec:dsm}, theoretical aspects of Doubly Stochastic Matrices are introduced and, afterward, in Section~\ref{sec:ls} learning and sampling procedures are explored. 
Section~\ref{sec:eda} presents the EDA algorithmic scheme adopted and discuss its characteristics.
Preliminary experiments are summarized in Section~\ref{sec:exp}. The paper concludes in Section~\ref{sec:concl} with a summary of the contribution and by proposing a number of lines for future research.

\section{Representations and Encodings in Permutation Problems}\label{sec:perm}

Formally, a permutation is a bijection function of the set of items $[n] = \{1,\ldots,n\}$ onto itself. Usually, the Greek letters $\sigma, \pi$ or $\rho$ are used to denote them. $\sigma(i)$ (also denoted as $\sigma_i$ for readability) represents the item at position $i$, and $\sigma^{-1}(i)$ represents the position of item $i$. The set of all permutations of size $n$ is denoted as $\mathbb{S}_n$, defines a group under the composition operation, and is known as the symmetric group in algebra.

The classical description of a permutation is the usual vector representation, where the items in the set $[n]$ appear in a particular order. Nevertheless, beyond an ordered set of items, a permutation can be represented uniquely (i.e., there exists a bijection between the vector representation and the following) by a collection of disjoint cycles, transpositions, pairwise precedence, 0/1 matrices, or even as graphs. Probably, due to the innumerable possibilities in which permutations can be encoded, they have served to represent solutions in combinatorial problems of different nature.

\begin{figure*}
    \centering
\includegraphics[width=\textwidth]{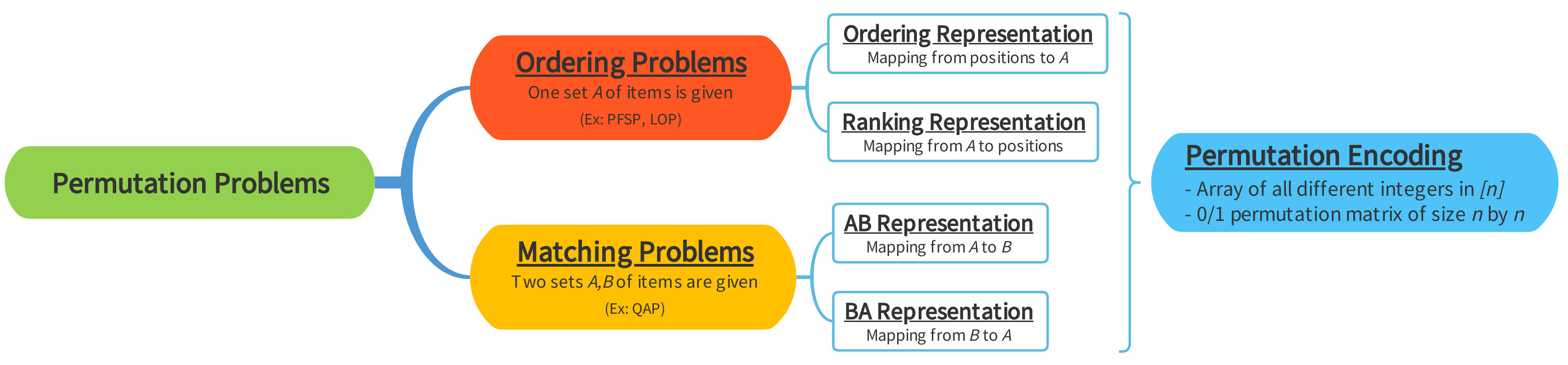}
    \caption{Relations among semantic interpretations, representations, and encodings in permutation problems.}
    \label{fig:tree}
\end{figure*}

Santucci et al.~\cite{Santucci2022} classified existing problems into two families (see also Fig.~\ref{fig:tree}):
\begin{itemize}
    \item \textit{ordering problems}, where the goal is to find an optimal ordering of a given set of items (as e.g., in the permutation flowshop scheduling problem), and
    \item \textit{matching problems}, where it is required to match, in the best possible way, two given equally sized sets of items (as e.g., in the quadratic assignment problem).
\end{itemize}

For both, ordering and matching problems, the fact that permutations are bijections of the first $n$ integers has to be intended only as a genotypic encoding. In an ordering problem, a set $A$ of $n$ items to be optimally ordered (on the basis of a provided objective function) is given.
Hence, an ordering of the items in $A$ can be represented in two distinct ways: as a mapping from positions to items (\textit{ordering representation}), or as a mapping from items to positions (\textit{ranking representation}).
Note that the semantic interpretation (ordering of items) is exactly the same for both the representations (ordering or ranking).
Once chosen a representation, the next step is to find a genotypic encoding.
Clearly, positions are integers in $[n]$, while the items in $A$ can be arbitrarily assigned to (all-different) identification numbers in $[n]$.
Therefore, both positions and items are encoded as elements of~$[n]$, thus a mapping between them can be easily encoded by a permutation in $\mathcal{S}_n$.
In other words, the ordering and ranking representations share the same genotypic encoding, but they still remain different representations of the same semantic interpretation. Note that in the case of ordering problems, under the ordering representation, the critical information about the solution is provided by the precedence between the items in the sequence (or the comparison of the magnitude of the rank under the ranking representation).

The genotype of matching problems describes a different idea. Specifically, the permutation describes a bijective function that matches the items in two discrete sets $A$ and $B$ of equal size $n$. Analogously to ordering problems, there are also two representations for matching problems: (1) {\it AB representation} and (2) {\it BA representation}. In the first case, each position in $[n]$ refers to the item from the set $A$, and the identification number placed in that position, denotes the item from $B$ to which the matching has been made. The BA representation, as the reader can guess, swaps the sets $A$ and $B$ from position and identification numbers.

In either problem, ordering or matching, it is possible to easily convert one representation to the other under the inversion operation. 

As stated in the introduction of this section, permutations can handle different descriptions beyond the usual vector representation. Two permutation encoding schemes which are relevant for this work are:
(1) arrays of $n$ all different integers from $[n]$ (vector representation), and
(2) $n$-by-$n$ permutation matrices, i.e., 0/1 matrices such that each row and each column has exactly one 1-entry (0/1 matrix representation).
The two encoding schemes are clearly equivalent, and they can be converted to each other by simple conversion procedures. In the following, we denote by $\mathbb{S}_n$ and $\mathbb{P}_n$ the sets of linearly encoded \mbox{$n$-length} permutations and $n$-by-$n$ permutation matrices, respectively. Hence, the conversion procedures realize an isomorphism between $\mathbb{S}_n$ and $\mathbb{P}_n$.

\section{Doubly Stochastic Matrices}\label{sec:dsm}


A Doubly Stochastic Matrix (DSM) is a matrix $D=[d_{ij}]_{n\times n}$ of non-negative real numbers such that each one of its rows and columns sums to 1, i.e., $d_{ij} \geq 0$ and $\sum_{i=1}^n d_{ij} = \sum_{j=1}^n d_{ij} = 1$, for all the items $i,j \in [n]$.
We denote by $\mathbb{D}_n$ the set of all the DSMs of order $n$.

It is easy to see that any permutation matrix is also a DSM but not vice versa, thus $\mathbb{P}_n \subset \mathbb{D}_n$.
In particular, according to the Birkhoff-von Neumann~(BvN) theorem~\cite{dufosse2016notes}, $\mathbb{D}_n$ defines a polytope which is the convex hull of $\mathbb{P}_n$.
As a consequence, any DSM can be written as a convex combination of permutation matrices.
Formally, given $D \in \mathbb{D}_n$, there exist $k$ permutation matrices $P_1, \ldots, P_k \in \mathbb{P}_n$ and $k$ weights $w_1, \ldots, w_k \in (0,1]$ summing up to 1 (i.e., $\sum_{i=1}^k w_i = 1$) such that
\begin{equation}
    D = w_1 P_1 + w_2 P_2 + \ldots + w_k P_k .
    \label{eq:decomp}
\end{equation}
The right-hand side of Eq.~(\ref{eq:decomp}) is said to be a decomposition of $D$ and it can be computed by the so-called Birkhoff algorithm~\cite{johnson1960algorithm}, which is known to return a decomposition whose length $k$ is at most $n^2 - 2n + 2$, i.e., $k=O(n^2)$.


By definition, any row and any column of a DSM is a multinomial\footnote{Also called "multinoully distribution" in~\cite{Goodfellow-et-al-2016}.} distribution over the set $[n]$ of row/column indices.
This property makes DSMs particularly appealing to build models for the AB (or BA) representation in the context of permutation matching problems.

As a prominent example of matching problems, let us consider the QAP, where it is required to match a set of $n$ facilities to a set of $n$ locations (i.e., the sets \textit{A} and \textit{B} of the AB representation).
Both the facilities and the locations can be encoded by the elements of $[n]$.
However, note that QAP, being a matching problem, does not define any ordering, neither in the set of facilities nor in the set of locations, thus the elements in $[n]$ --though being cardinal numbers-- have to be intended as identification numbers without relying on any ordering relation among them.
In this context, we can design a DSM $D \in \mathbb{D}_n$ such that:
\begin{itemize}
    \item the rows and columns of $D$ represent, respectively, facilities and locations;
    \item the $i$-th row of $D$ is a multinomial distribution representing the probabilities of assigning any location to the facility $i$;
    \item the $j$-th column of $D$ is a multinomial distribution representing the probabilities of assigning any facility to the location $j$.
\end{itemize}
The points above clearly show that DSMs are suitable probability models for matching problems, being able to coherently encode both facilities and locations' distributions without relying on any ordering relation among them.

\section {Learning and Sampling DSMs}\label{sec:ls}

The BvN theorem previously presented clearly states that a DSM can be seen as the aggregation of a number of permutations (encoded in the form of permutation matrices) such that only the matching or assignment nature of the permutations is considered.
This property makes particularly appealing the use of DSMs as models for permutation matching problems.
Therefore, here below we analyze different methodologies for learning a DSM from a set of permutations and, vice versa, sampling permutations from a DSM.

\subsection{Learning DSMs}\label{sec:dsm_learning}

In order to learn a DSM from a set of permutations we devise two different strategies: \textit{exact learning} and \textit{smoothed learning}, of which the latter is a generalization of the former.

\vspace{2mm} \noindent \textbf{Exact Learning.}
Given $m$ permutations $P_1,\dots,P_m \in \mathbb{P}_n$, then a DSM $D \in \mathbb{D}_n$ can be learned as a convex combination of them.
Formally, 
\begin{equation}
    D = w_1 P_1 + w_2 P_2 + \ldots + w_m P_m ,
    \label{eq:learn1}
\end{equation}
where $w_1,\ldots,w_m$ are $m$ non-negative weights summing to 1, that can be uniformly set to $1/m$ or, alternatively, they can be made proportional to some importance measure of the corresponding permutations.

According to the BvN theorem (see Sect.~\ref{sec:dsm}), $D$ can also be interpreted as the (weighted) centroid of $P_1,\ldots,P_n$ in the Birkhoff polytope.

As a simple consequence of the convex combination, if the generic entry $(i,j)$ is null in all the permutation matrices $P_1,\ldots,P_m$, then also the entry $(i,j)$ of $D$ is null, and vice versa.
As it will be clear from the next subsection, this property makes it impossible to sample from $D$, permutations which assign the facility~$i$ to the location $j$.

\vspace{2mm} \noindent \textbf{Smoothed Learning.}
In order to address the "null probability" drawback of the \textit{exact learning} scheme, we slightly modify Eq.~(\ref{eq:learn1}) by introducing --as a further term of the convex combination-- the uniform DSM $U = [u_{ij}]_{n \times n}$ such that $u_{ij} = 1/n$ for every $i,j \in [n]$.
Formally, a DSM $D$ can now be learned as follows.
\begin{equation}
    D = w_1 P_1 + w_2 P_2 + \ldots + w_m P_m + \alpha U ,
    \label{eq:learn2}
\end{equation}
where $\alpha \in (0,1]$ is a smoothing factor which regulates the importance of $U$ in the convex combination, while now $\sum_{i=1}^m w_i = 1-\alpha$.
Therefore, in the uniform setting, the weights are set to $(1-\alpha)/m$.

The learning scheme of Eq.~(\ref{eq:learn2}) produces a DSM without 0-entries, thus making possible to sample permutations containing items' assignments not present in the training set.
Since this aspect is relevant for avoiding the premature convergence of an estimation-of-distribution algorithm based on the DSM model, in this work we will adopt the \textit{smoothed learning} scheme.

\subsection{Sampling DSMs}\label{sec:dsm_sampling}

In order to sample permutations from a given DSM we devise three different strategies:
\begin{itemize}
    \item \textit{probabilistic sampling}, which allows to define a proper probability mass function;
    \item \textit{algebraic sampling}, that exploits the multiplication of the DSM by a random vector;
    \item \textit{geometric sampling}, which is based on the BvN theorem.
\end{itemize}
These strategies are described in the following.

\vspace{2mm} \noindent \textbf{Probabilistic Sampling (PS).}
For the sake of simplicity, in order to describe this sampling procedure, we use the linear permutation encoding.
Therefore, given $D \in \mathbb{D}_n$, a permutation $\sigma \in \mathbb{S}_n$ is sampled from $D$ as follows.

\begin{enumerate}
    \item Select uniformly at random a row or a column of $D$ and denote by $i \in [n]$ and $p \in [0,1]^n$ its index and the corresponding probability vector, respectively.
    \item Since $p$ is a multinomial probability distribution, sample an item $j \in [n]$ according to $p$.
    \item If $p$ is a row of $D$, then set $\sigma_i := j$, otherwise set $\sigma_j := i$.
    \item Remove the row $i$ (or $j$) and the column $j$ (or $i$) of $D$ when $p$ is a row (or a column), then renormalize the remaining rows and columns.
    \item If not all the rows and columns of $D$ have been removed, go back to step (1).
\end{enumerate}

It is easy to see that any iteration of this procedure sets exactly one entry of $\sigma$ and removes exactly one row and one column from~$D$.
This guarantees that, after $n$ iterations, $\sigma$ is a proper permutation of $\mathbb{S}_n$.
Moreover, since the computational cost of a single iteration is~$\Theta(n)$, sampling a permutation with the PS costs $\Theta(n^2)$ time steps.

For a given $n$-by-$n$ DSM $D=[d_{ij}]$, the probability of sampling a permutation $\sigma \in \mathbb{S}_n$, via PS, is given by
\begin{equation}
    \mbox{Pr}(\sigma|D) = 
    \frac{ \prod_{i=1}^n d_{i,\sigma_i} }
         { \sum_{\pi \in \mathbb{S}_n} \prod_{i=1}^n d_{i,\pi_i}  }
         =
    \frac{ \prod_{i=1}^n d_{i,\sigma_i} }
         { \mbox{Perm}(D) }
    ,
    \label{eq:prob}
\end{equation}
where the denominator is the permanent of the matrix $D$~\cite{glynn2010permanent}.

The probability mass function of Eq. (\ref{eq:prob}) allows to derive two interesting properties of PS as follows.
By recalling that permutation matrices are particular cases of DSMs, when the current DSM is a permutation matrix $P$, then only $P$ can be sampled.
At the other extreme, when the current DSM is the uniform DSM $U$ (introduced in Sect.~\ref{sec:dsm_learning}), all the permutations are equiprobable.

\vspace{2mm} \noindent \textbf{Algebraic Sampling (AS).}
This sampling strategy is based on the "randomized rounding" methodology introduced in~\cite{wolstenholme2016sampling} (which, in turn, extends the methods described in~\cite{fogel2013convex} and~\cite{barvinok2006approximating}).

The idea is that, to sample a permutation matrix $P \in \mathbb{P}_n$ from a $D \in \mathbb{D}_n$, we first generate a vector $v \in [0,1]^n$ uniformly at random and then we obtain $P$ by solving the equation
\begin{equation}
    P \cdot \mbox{rank}(v) = \mbox{rank}(D \cdot v) ,
    \label{eq:randomizedrounding}
\end{equation}
where $\cdot$ is the usual matrix-vector multiplication, while the vector $\mbox{rank}(v)$ is defined as \mbox{$\mbox{rank}(v)_i = j$}, where $v_i$ is the $j$-th smallest value in $v$.
For example, if $v=(0.1,0.5,0.8,0.2)$, then $\mbox{rank}(v)=(1,3,4,2)$.
Practically, $\mbox{rank}(v)$ returns the inverse permutation of $\mbox{argsort}(v)$, as the well-known random key transformation for permutations~\cite{santucci2020algebraic}.

Therefore, by using the linear permutation encoding, a permutation $\sigma \in \mathbb{S}_n$ is sampled from a DSM $D \in \mathbb{D}_n$ as follows.
\begin{enumerate}
    \item Generate a vector $v \in [0,1]^n$ uniformly at random.
    \item Calculate the permutations $\pi = \mbox{rank}(D \cdot v)$ and $\rho = \mbox{argsort}(v)$.
    \item Returns the permutation $\sigma = \pi \circ \rho$ (where $\circ$ is the usual permutation composition).
\end{enumerate}

In~\cite{wolstenholme2016sampling} it has been proved that the linear permutation $\sigma$, computed as above, corresponds to the permutation matrix $P^*$ which solves the following optimization problem:
\begin{equation}
    P^* = \mbox{argmin}_{P \in \mathbb{P}_n} ||D \cdot v - P \cdot v||_F^{2} ,
    \label{eq:optprobl}
\end{equation}
where $||\cdot||_F$ is the usual Frobenius norm of a matrix.

Eq.~(\ref{eq:optprobl}) provides an intuitive justification to AS.
Moreover, it also shows that, as for the PS case, when $v$ entries are all different\footnote{Since $v$ entries are sampled independently and uniformly at random, they are all different with probability 1.} and the current DSM is a permutation matrix, then only this permutation can be sampled.
Conversely, when the DSM is the uniform one, all the permutations have the same probability to be sampled.

Finally, the computational complexity of the procedure is $\Theta(n^2)$, which is given by the matrix-vector multiplication in step (2).
However, it is worthwhile to note that the matrix-vector multiplication can benefit of very fast implementations, thus making AS very efficient in practice.

\vspace{2mm} \noindent \textbf{Geometric Sampling (GS).}
Given a DSM $D \in \mathbb{D}_n$, it is possible to sample a permutation $P \in \mathbb{P}_n$ --expressed in the form of a permutation matrix for convenience of description-- by means of the following steps.
\begin{enumerate}
    \item Execute the Birkhoff algorithm~\cite{johnson1960algorithm} and obtain a decomposition of $D$ with length $k = O(n^2)$.
    \item The decomposition has the form of Eq.~(\ref{eq:decomp}), hence sample an index $i \in [k]$ according to the decomposition weights $w_1,\ldots,w_k$.
    \item Select and return the permutation matrix $P_i$ from the computed decomposition.
\end{enumerate}

Substantially, the returned permutation is sampled from the vertices of the Birkhoff polytope by implicitly considering probabilities which are inversely proportional to the geometric distances of each vertex from the point corresponding to $D$.

As in the PS and AS cases, when the DSM is a permutation matrix, its Birkhoff decomposition is formed only by itself, thus it is the only permutation that can be sampled.
However, when the DSM is the uniform DSM, since the Birkhoff algorithm is not randomized (or, at least, not yet), not all the permutations are equiprobable.
In fact, an easy to see drawback of GS is that the sampling domain is limited by BvN theorem, i.e., no more than $n^2-2n+2$ different permutations can be sampled.

Furthermore, GS is not efficient at all.
Indeed, the computational complexity of GS is dominated by the Birkhoff algorithm which, as described in~\cite{johnson1960algorithm}, is a greedy iterative method that, at any iteration, calculates a perfect matching in a bipartite graph formed by $2n$ vertices.
The most used solver for bipartite matching is the Hopcroft-Karp algorithm~\cite{hopcroft1973n}, whose computational complexity is $O(n^{2.5})$.
Note also that, due to the BvN theorem, the Birkhoff algorithm requires $O(n^2)$ iterations, therefore GS has the very high computational complexity of $O(n^{4.5})$.

\section{The proposed EDA}\label{sec:eda}

In this section we introduce an Estimation of Distribution (EDA) for permutation matching problems whose probability model is designed as a DSM.

We adopt the \textit{smoothed learning} scheme introduced in Sect.~\ref{sec:dsm_learning} and the three sampling strategies introduced in Sect.~\ref{sec:dsm_sampling}.
Therefore, we have three different implementations of our EDA to which we refer with the terms DSM-PS, DSM-AS and DSM-GS, on the basis of the sampling scheme adopted.

Apart from the sampling strategy, all the implementations share the same base algorithm whose pseudocode is provided in Alg.~\ref{alg:dsm}.

\begin{algorithm}[!htbp]
    \caption{Algorithmic scheme of the proposed EDA}
    \renewcommand{\algorithmicrequire}{\textbf{Input:}}
    \renewcommand{\algorithmicensure}{\textbf{Output:}}
    \begin{algorithmic}[1]\small
        \Require $f: \mathbb{S}_n \to \mathbb{R}$, \, $\mathrm{SS} \in \{ \mathrm{PS}, \mathrm{AS}, \mathrm{GS} \}$
        \State $\lambda \gets 10n$ \Comment{Sample size}
        \State $\mu \gets n$ \Comment{Selection size}
        \State $\alpha \gets 1/{n^2}$ \Comment{Smoothing factor}
        \State $t \gets 0$ 
        \State $X_0 \gets $ a set of $\lambda$ permutations drawn uniformly at random
        \While{the budget of evaluations is not exhausted}
            \State $t \gets t+1$
            \State $Y \gets $ select the best $\mu$ permutations from $X_{t-1}$
            \State $D \gets $ a DSM learnt from $Y$ with smoothing factor $\alpha$
            \State $Z \gets $ a set of $\lambda$ permutations sampled from $D$ using $\mathrm{SS}$
            \State evaluate $f(\sigma)$ for all $\sigma \in Z$
            \State $X_t \gets Y \cup Z$
        \EndWhile
        \State \Return the best permutation in $X_t$
    \end{algorithmic}
    \label{alg:dsm}
\end{algorithm}

Alg.~\ref{alg:dsm} receives in input the objective function to optimize and the desired sampling strategy to be chosen among PS, AS, or GS.

In lines~1--3, three algorithmic parameters are set on the basis of the problem size $n$ and according to previous experiences in the field of EDAs~\cite{ceberio2013a,irurozki2014e}.
Namely, a high selection pressure is considered by setting the sample size $\lambda$ and the selection size $\mu$ to, respectively, $10n$ and $n$.
The smoothing factor $\alpha$ is set to $1/n^2$ in order to make possible to escape stagnation states, though without making the search totally random.

Then, a set $X$ of solutions is initialized with $\lambda$ random permutations in line 5 and iteratively updated in the main loop of lines~6--13.
At any iteration of the loop, the best $\mu$ solutions in $X$ (line~8) are used to learn the DSM model $D$ (line~9), from which $\lambda$ new solutions are sampled (line~10).
The sampled solutions are evaluated (line~11) and $X$ is updated by merging the previously selected solutions with the new generated ones (line~12).

Note that, in the learning step, apart from $\alpha$, all the weights for the solutions in the training set are set in a uniform way.

Moreover, the iterative update scheme ensures an elitist behaviour to the algorithm.
Indeed, the set $X$ always contains the best solution ever visited, which is finally returned in line~14.
Furthermore, the elitism guarantees that, at any iteration $t$, the DSM model $D$ is always learnt from the best $\mu$ solutions ever visited by the algorithm.

The elitist behaviour and the high selection pressure are counterbalanced by the smoothed learning which allows to generate unseen items' matchings with a non-null probability.

As discussed, learning and sampling are carried out by using, respectively, the smoothed learning scheme and the chosen sampling strategy.
Therefore, the complexity of an iteration is dominated by the sampling and evaluation steps.




\section{Experimental Study}\label{sec:exp}

In order to analyze the effectiveness and efficiency of the explored solutions for the sampling and learning methods of DSMs, in the following we present a thorough experimental study.

\subsection{Experimental setting}
As exposed in the introduction, the present work is oriented to matching/assignment type problems and, thus, for the purpose of the experimentation, we have chosen the Quadratic Assignment Problem (QAP) as the benchmark problem. In the QAP, we are given two square matrices $\mathbf{B}=[b_{ij}]_{n\times n}$ and $\mathbf{H}=[h_{ij}]_{n\times n}$ of parameters and the goal is to find the permutation $\sigma\in\mathbb{S}_n$ that minimizes the objective function \[f(\sigma)=\sum_{i=1}^{n}\sum_{j=1}^{n}b_{i,j} h_{\sigma(i)\sigma(j)}.\]

Twelve instances of a variety of sizes from the QAPLIB~\cite{burkard1997} have been chosen to illustrate the experiments, and the selection of the instances was made prior to observing any performance results.
In particular, the set of selected instances is formed by: {\it tai15\{a/b\}, tai20\{a/b\}, tai30\{a/b\}, tai50\{a/b\}, tai80\{a/b\}} and {\it tai100\{a/b\}}. 

In order to evaluate the usability of DSMs with optimization purposes, the three sampling procedures described in Sect.~\ref{sec:dsm_sampling} \mbox{--namely, PS, AS, and GS--} are considered. To have a real view of their performance, we will also consider three other competitor EDA algorithms from the literature:
\begin{itemize}
\item Mallows EDA under the Cayley distance~\cite{irurozki2018sampling}. This is the main competitor of the proposed algorithms in this work.
\item Mallows EDA under the Kendall distance~\cite{ceberio2013a}. Due to the nature of the metric employed, this algorithm should not be suited to deal with matching problems.
\item Plackett-Luce EDA~\cite{ceberio2013plackett} has been acknowledged for being an algorithm to deal with ordering problems, and thus, as the previous, should not be very effective in matching problems.
\end{itemize}
The labels used to denote them are EDA-MC, EDA-MK and EDA-PL, respectively. Similarly, we distinguish among the DSM algorithms, as DSM-PS and DSM-AS with respect to the sampling method.

Regarding the geometric sampling, preliminary runs point out that DSM-GS the method is not practicable in this framework.
In fact, it was observed that the GS sampling strategy mostly produces the same permutations which have been used to learn the DSM, even with very large values for the smoothed factor $\alpha$. This in turn makes DSM-GS produce very rare improvements in the best-so-far solution. Therefore, by also considering the very high computational complexity of GS (see Sect.~\ref{sec:dsm_sampling}), we decided to omit it in the presentation of the experimental results, deferring a more accurate analysis to future work.

A general parameter settings has been decided for all the algorithms, following the guidelines of the EDA competitors in the literature. 
Hence, as also discussed in Sect.~\ref{sec:eda}, the sample size and selection size are set to $\lambda=10n$ and $\mu=n$, respectively.

\subsection{Effectiveness}

In order to obtain a general view of the effectiveness of the DSM type algorithms when compared to the EDAs mentioned above, we executed each algorithm for 20 repetitions on the selected set of benchmark instances, while a budget of $100n^2$ evaluations is allowed for each execution. Results are summarized in Table~\ref{tab:performance100n2} as Median Relative Deviations (MRD) with respect to the best known results reported in the QAPLIB website\footnote{https://coral.ise.lehigh.edu/data-sets/qaplib/qaplib-problem-instances-and-solutions/}. Surprisingly, results show that there is one algorithm that obtains much better results than the rest of the algorithms: DSM-PS. The rest of the algorithms seem to be worse than DSM-PS and have similar performance among them.

\begin{table}[h]
    \centering
    \small
    \caption{Results of the EDA and DSM algorithms for the 12 QAP instances from the QAPLIB benchmark (from the Taillard set). The Median Relative Deviations (MRD) measures of the values found across the 20 repetitions of the best known results are reported. Results in bold highlight the algorithm that obtained the lowest MRD. A maximum number of $100n^2$ evaluations were performed by each of the algorithms.}
    \resizebox{\linewidth}{!}{
    \begin{tabular}{l|r||rrr|rr}
Instance & Best Known & EDA-PL & EDA-MK & EDA-MC & DSM-PS &  DSM-AS\\ \hline
tai15a & 388214 & 0.09283 & 0.09039 & 0.06565 & {\bf 0.04145} &  0.09659\\
tai15b & 51765268 & 0.01379 & 0.00960 & 0.00510 & {\bf 0.00455} &  0.01520\\
tai20a & 703482 & 0.13623 & 0.13740 & 0.13293 & {\bf 0.05471} &  0.13805\\
tai20b & 122455319 & 0.11341 & 0.05344 & 0.04272 & {\bf 0.01807} &  0.11903\\
tai30a & 1818146 & 0.12511 & 0.12666 & 0.13055 & {\bf 0.04783} &  0.13016\\
tai30b & 637117113 & 0.27583 & 0.15849 & 0.31548 & {\bf 0.07563} &  0.31814\\
tai50a & 4938796 & 0.14096 & 0.14025 & 0.13820 & {\bf 0.04965} &  0.14048\\
tai50b & 458821517 & 0.35262 & 0.09779 & 0.36341 & {\bf 0.03915} &  0.37153\\
tai80a & 13499184 & 0.12401 & 0.12492 & 0.12404 & {\bf 0.03945} &  0.12495\\
tai80b & 818415043 & 0.34160 & 0.31773 & 0.34531 & {\bf 0.03729} &  0.34593\\
tai100a & 21052466 & 0.11712 & 0.11668 & 0.11664 & {\bf 0.03480} &  0.11713\\
tai100b & 1185996137 & 0.32555 & 0.12570 & 0.33081 & {\bf 0.01785} &  0.33082\\
    \end{tabular}
    }
    \label{tab:performance100n2}
\end{table}

\begin{figure}
    \centering
    \includegraphics[width=8cm]{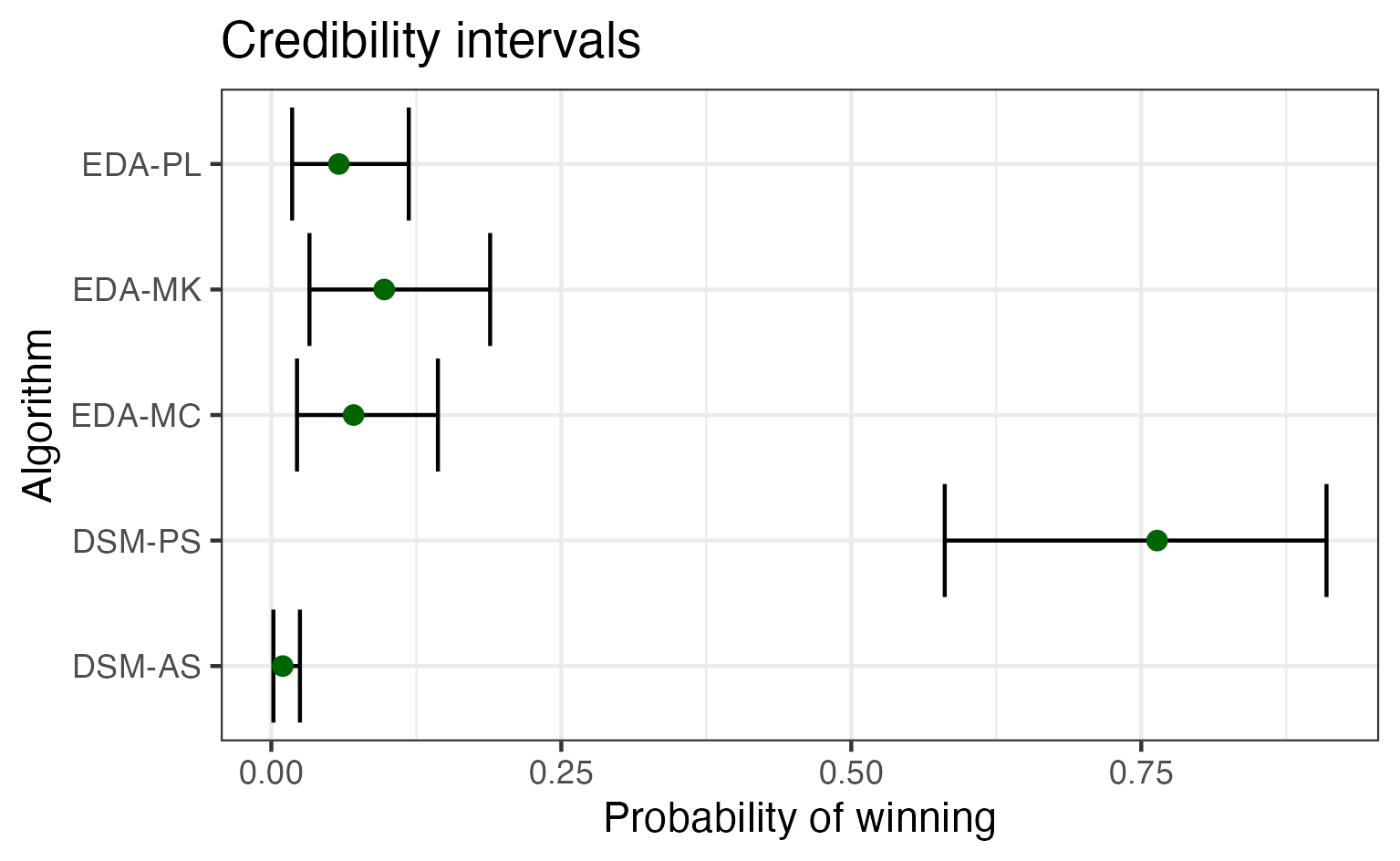}
    \caption{Credibility intervals of the evaluated algorithms on the set of benchmark instances. The intervals describe for each algorithm the probability of being the best option, based on the experimental data provided.}
    \label{fig:bayesian}
\end{figure}

To statistically assess the differences among the different alternatives (combination of parameters), a Bayesian performance analysis was carried out~\cite{Rojas2022}. The outcome of the analysis is summarized in Fig.~\ref{fig:bayesian} in the form of credibility intervals.

In the $y$-axis of the plot the different algorithms are listed, and for each case, a credibility interval is depicted. Such interval, formed with a green dot (the expectancy) and a range of values, describes the probability of that algorithm being the best alternative among the compared ones. The values needed to build the intervals are obtained by sampling the posterior distribution of the Bayesian model computed\footnote{A detailed procedure for the analysis is provided in~\cite{Rojas2022}.}.

The analysis confirms that DSM-PS is one step ahead of the other algorithms, however, the width of the intervals, especially for EDA-PL and DSM-PS suggests that there is still much uncertainty related to previous statement. Taking into account that only 12 instances were used for the analysis, it becomes obvious that additional data is required to have more reliable conclusions. However, this results already point that DSMs may be very valuable for optimization purposes, at least in the context of matching problems.

\subsection{Efficiency}

We have seen that DSMs can offer superior performance to some competitors in the literature, but, what about their computational efficiency? In this section, we aim to evaluate the computational time required by the algorithms to run $100n^2$ evaluations and how it scales with the problem size. At this point, it is important to remark that EDAs were implemented in C++ while, and DSM versions were coded in Python, so this aspect needs to be taken into account when doing the analysis\footnote{A comparison of the energy efficiency of 27 programming languages is carried out in~\cite{pereira2021ranking}. The analysis reveals that while C and C++ are at the top of the ranking, Python usually is the one of the worst ranked, if not the worst.}.

\begin{figure}
  \centering
    \includegraphics[width=8cm]{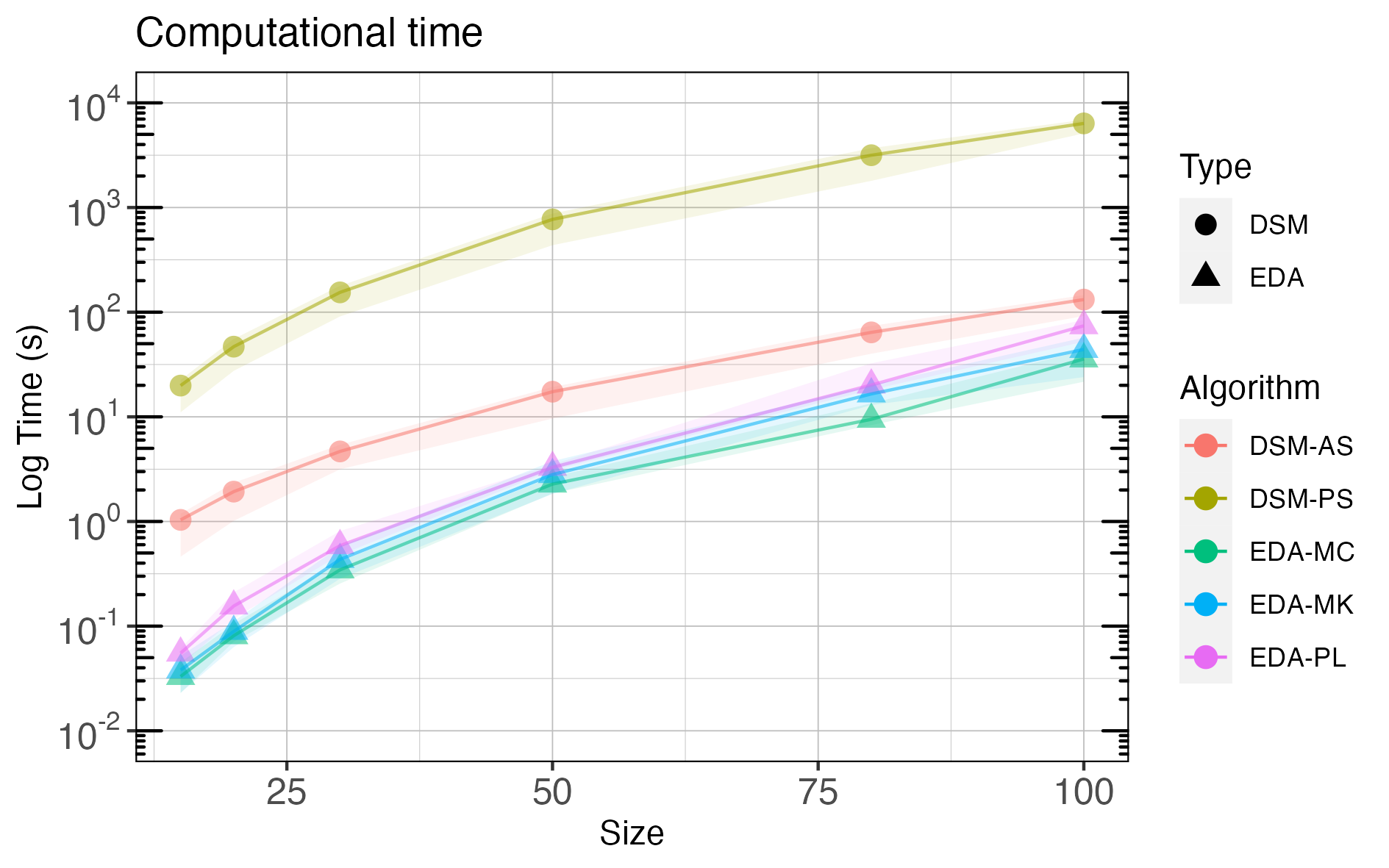}
    \caption{Average execution time (log seconds) of each run of the EDA-PL, EDA-MK, EDA-MC, and DSM-AS, DSM-PS algorithms for running $100n^2$ evaluations.}
    \label{fig:execution_time}
\end{figure}

Average results of the computational time consumed by each of the compared algorithms are depicted in Figure~\ref{fig:execution_time}. As the number of evaluations to perform is a function of the problem size, we have plotted a line for each algorithm for the different problem sizes considered in the instance set. Results reveal that EDAs (all of them implemented in C++), have much lower computational cost compared to the proposed DSM-based algorithms (note that $y$-axis is in log scale). In fact, the best performing algorithm in the previous section, DSM-PS, is by far the most time-consuming method. The DSM algorithms are prototype versions that have been implemented in Python and the room for improvement is high, and therefore the chances of approaching the performance of EDAs is real.

This last statement can be further substantiated by the fact that Figure~\ref{fig:execution_time} shows that the scalability of the algorithms is similar, as the slopes of the line plots are comparable.

\subsection{Sampling convergence}

As a way to analyze how effective are the different algorithms when modelling and capturing the relevant information of the candidate solutions for the problem, in this section, we analyze the quality of the solutions sampled at each iteration of the algorithm. We do not focus on the population itself, since its quality is usually monotonically increasing, but we aim to see the solutions sampled. In an appropriate model, we expect to see that the quality of the sampled solutions improves across the iterations.

\begin{figure}
  \centering
    \includegraphics[width=8cm]{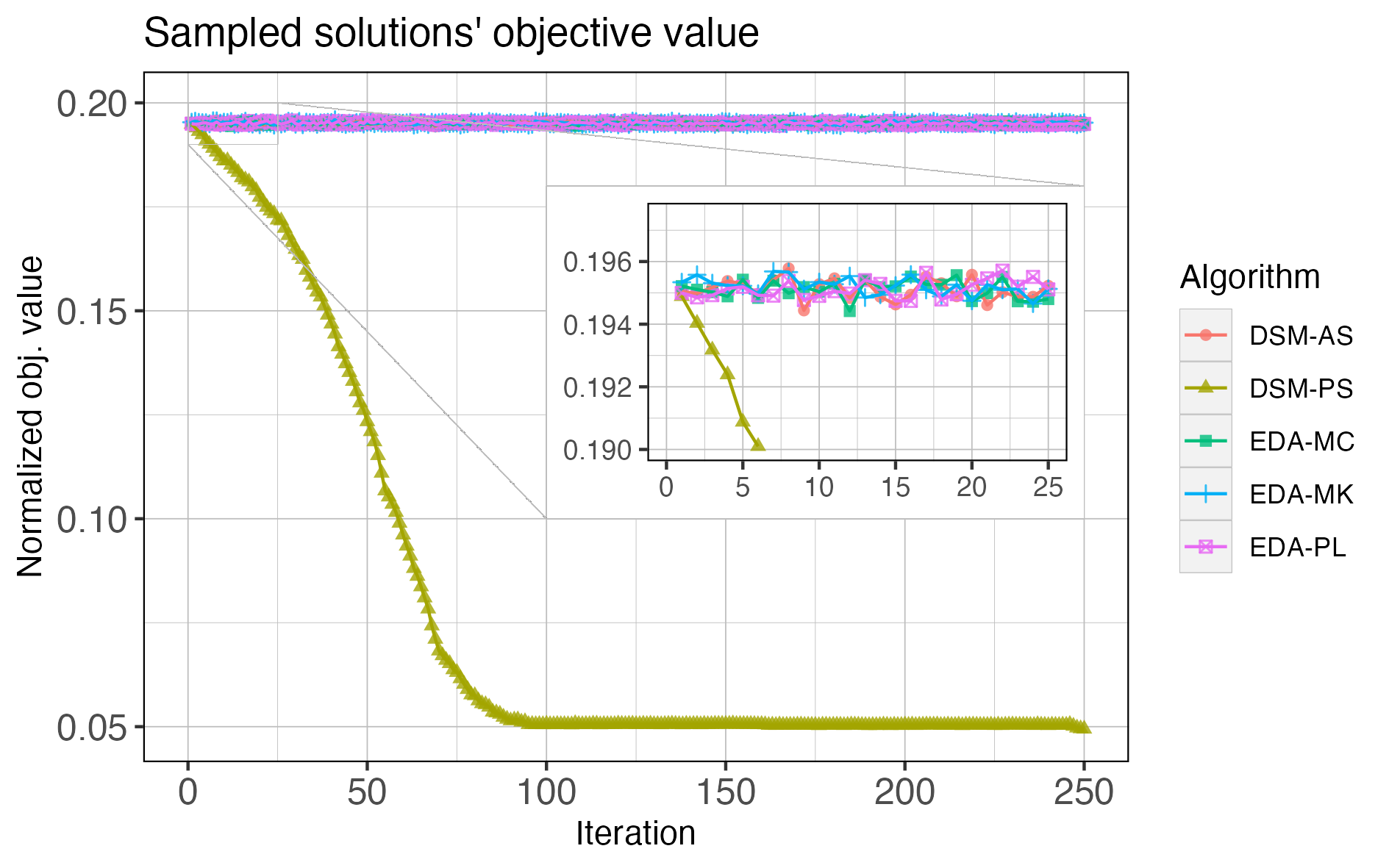}
    \caption{Average objective function value of the solutions sampled at each iteration of EDA-PL, EDA-MK, EDA-MC, and DSM-AS, DSM-PS algorithms for running $100n^2$ evaluations. Each algorithm was run 5 repetitions. The instance used in this plot, is {\it tai50a}. $x$-axis has been limited to 250 iterations.}
    \label{fig:convergence}
\end{figure}

To that end, we have conducted an alternative experiment on the instance {\it tai50a} where every solution sampled during the execution by each algorithm has been recorded. Then, the objective value of the samples at each iteration is averaged, and to avoid randomness, 5 repetitions for each algorithm were performed. Results are presented in the form of scatter and smooth line plots in Figure~\ref{fig:convergence}.

On the one hand, we can observe that the EDAs show almost no convergence, and they do not move from (normalized) objective values within the range $[0.19,0.2]$. With respect to the DSMs, the results are very variable. DSM-PS shows a very particular shape where we see that almost since the first iterations, the algorithm starts progressively improving the results. This cannot be generalized to DSMs, as the behaviour of DSM-AS seems equivalent to other EDAs.

\section{Conclusion and Future Work}\label{sec:concl}

In this paper, we investigated the use of Doubly Stochastic Matrices (DSMs) in the framework of evolutionary algorithms and focused on Estimation of Distribution Algorithms (EDAs) for permutation-based combinatorial optimization problems.

We consider that existing EDAs are not suited for solving matching or assignment-type permutation problems, as most of them assume an ordering nature of the permutations. In this context, we think that DSMs can have good performance. To that end, we analyzed the potential use of DSMs within EDAs, by exploring different learning and sampling methods. In particular, we designed a simple learning process by exploiting the Birkhoff-von Neumann theorem, which is a well known  result in the algebraic field. Regarding the sampling of candidate solutions (permutations) from a DSM, we found multiple ways to do so. In this paper, we studied three sampling strategies --probabilistic, algebraic, and geometric-- with characteristics and variable time complexities.

The conducted experiments on a set of instances of the quadratic assignment problem reveal an interesting scenario. When compared with already published EDA algorithms for permutations problems, we observe that our DSM-based EDAs, particularly, under the probabilistic sampling, obtain very good, and even better, results. 

This work intends to explore the potential of doubly stochastic matrices within the framework of estimation of distribution algorithms, and thus, there is a lot to improve in their design, integration, and fine-tuning. On the one hand, the computational time required by these algorithms needs still further research to make them more competitive. The fact that the DSM-based algorithms were coded in Python, while the competitor EDAs were implemented in C++, suggests that the efficiency gap observed can be narrowed down. On the other hand, DSM-based algorithms can be further refined by conducting a deeper analysis of the parameters' setting (now hard-coded in the algorithm) and by also studying a novel weighting scheme (now uniform) for the learning phase.
Moreover, it is also interesting to explore other sampling methodologies, such as the one based on the Sinkhorn-Knopp algorithm which has been recently adopted in the field of machine learning \cite{mena2017sinkhorn}.

Finally, in this paper, we exclusively focused on implementing EDAs, as it was straightforward, but doubly stochastic matrices can have other uses such as, in the context of genetic algorithms, for designing probabilistic crossover operators tailored to permutation matching problems or, in the context of model-based gradient search algorithms, for designing differentiable Mallows-like models where the mode permutation is relaxed to a doubly stochastic matrix.

\section*{Acknowledgments}
Josu Ceberio has been partially supported by the Research Groups 2022-2025 (IT1504-22), and Elkartek (KK- 2021/00065, KK-2022/00106) from the Basque Government and the PID2019-106453GA-I00 research project from the Spanish Ministry of Economy, Industry and Competitiveness.
Valentino Santucci has been partially supported by the research projects: ``Universit\`{a} per Stranieri di Perugia -- Finanziamento per Progetti di Ricerca di Ateneo - PRA 2022'', ``Universit\`{a} per Stranieri di Perugia -- Artificial intelligence for education, social and human sciences'', and ``Universit\`{a} per Stranieri di Perugia -- Progettazione e sviluppo di strumenti digitali per la formazione a distanza''.

\bibliographystyle{unsrt}  
\bibliography{references}

\end{document}